\documentclass[letterpaper, 10 pt, conference]{ieeeconf}  

\IEEEoverridecommandlockouts                              

\usepackage[pdftex]{graphicx}
\graphicspath{{figures/}{../jpeg/}}
\usepackage{url}
\usepackage{bm}
\usepackage{amsfonts}
\usepackage{amssymb}
\usepackage{empheq}
\usepackage{color}  
\usepackage{caption}
\usepackage{subcaption}
\usepackage{booktabs}
\usepackage{comment}

\title{\LARGE \bf
Fast, On-board, Model-aided Visual-Inertial Odometry System for Quadrotor Micro Aerial Vehicles*
}

\author{Dinuka Abeywardena, Shoudong Huang, Ben Barnes, Gamini Dissanayake and Sarath Kodagoda
\thanks{*This work was supported by Centre for Autonomous Systems, University of Technology, Sydney.}
\thanks{All Authors are with the Centre for Autonomous Systems, University of Technology, Sydney.
        {\tt\small {Ben.Barnes@student.uts.edu.au, \{Dinuka.Abeywardena,Shoudong.Huang, Gamini.Dissanayake,Sarath.Kodagoda\}@uts.edu.au}}}%
}

\begin{document}

\maketitle
\thispagestyle{empty}
\pagestyle{empty}

\begin{abstract}
The main contribution of this paper is a high frequency, low-complexity, on-board visual-inertial odometry system for quadrotor micro air vehicles. The system consists of an extended Kalman filter (EKF) based state estimation algorithm that fuses information from a low cost MEMS inertial measurement unit acquired at 200Hz and VGA resolution images from a monocular camera at 50Hz. The dynamic model describing the quadrotor motion is employed in the estimation algorithm as a third source of information. Visual information is incorporated into the EKF by enforcing the epipolar constraint on features tracked between image pairs, avoiding the need to explicitly estimate the location of the tracked environmental features. Combined use of the dynamic model and epipolar constraints makes it possible to obtain drift free velocity and attitude estimates in the presence of both accelerometer and gyroscope biases. A strategy
to deal with the unobservability that arises when the quadrotor is in hover is also provided. Experimental data from a real-time implementation of the system on a 50 gram embedded computer are presented in addition to the simulations to demonstrate the efficacy of the proposed system. 
\end{abstract}

\section{INTRODUCTION}
The main task of a state estimator for a robotic platform is to produce fast, accurate and timely trajectory estimates to facilitate closed loop control. For robots with slow and inherently stable dynamics, infrequent and delayed state estimates may suffice to achieve satisfactory levels of control. However, fast and unstable dynamics of small unmanned aerial systems such as quadrotor micro aerial vehicles (MAV) require estimation algorithms capable of producing fast, real-time state estimates, if they are to be manoeuvred in a manner that fully exploits the extreme agility of such platforms.

It is common to envision a layered approach to the combined problem of estimation and control of MAV platforms with fast dynamics \cite{mahony2013}. The low level attitude rate, attitude and linear velocity estimators (and controllers) are tasked with the instantaneous stability of the platform and need to react as fast as possible to changes in true linear and angular positions. As the stability is ensured through the low level estimators and controllers, the position and heading estimators may need not be as fast, thereby forming a separate layer of a multi-rate estimation and control architecture.

Estimators based on GPS \cite{kim2006real} or Simultaneous Localization And Mapping (SLAM) algorithms \cite{nutzi2011} are suitable for the latter but not the former layers mentioned above as their estimates are infrequent and often delayed. In contrast, Inertial Measurement Units (IMU) can be sampled at rates of up to several hundred samples per second with negligible delay. Conventionally, algorithms that employ IMU measurements have formed the foundation of low level, real-time attitude estimators and controllers for MAVs \cite{mahony2008}. However, all such formulations that rely solely on inertial measurements for attitude estimation assume that the accelerometer biases are known and that vehicle inertial accelerations are small. Even under these assumptions, inertial sensors alone are not capable of providing drift free linear velocity estimates of MAVs due to various noise sources present in low cost MEMS accelerometers and gyroscopes.

%
\begin{figure}[tb]
\centering
\begin{subfigure}[b]{0.23\textwidth}
\centering
\includegraphics[width=\textwidth]{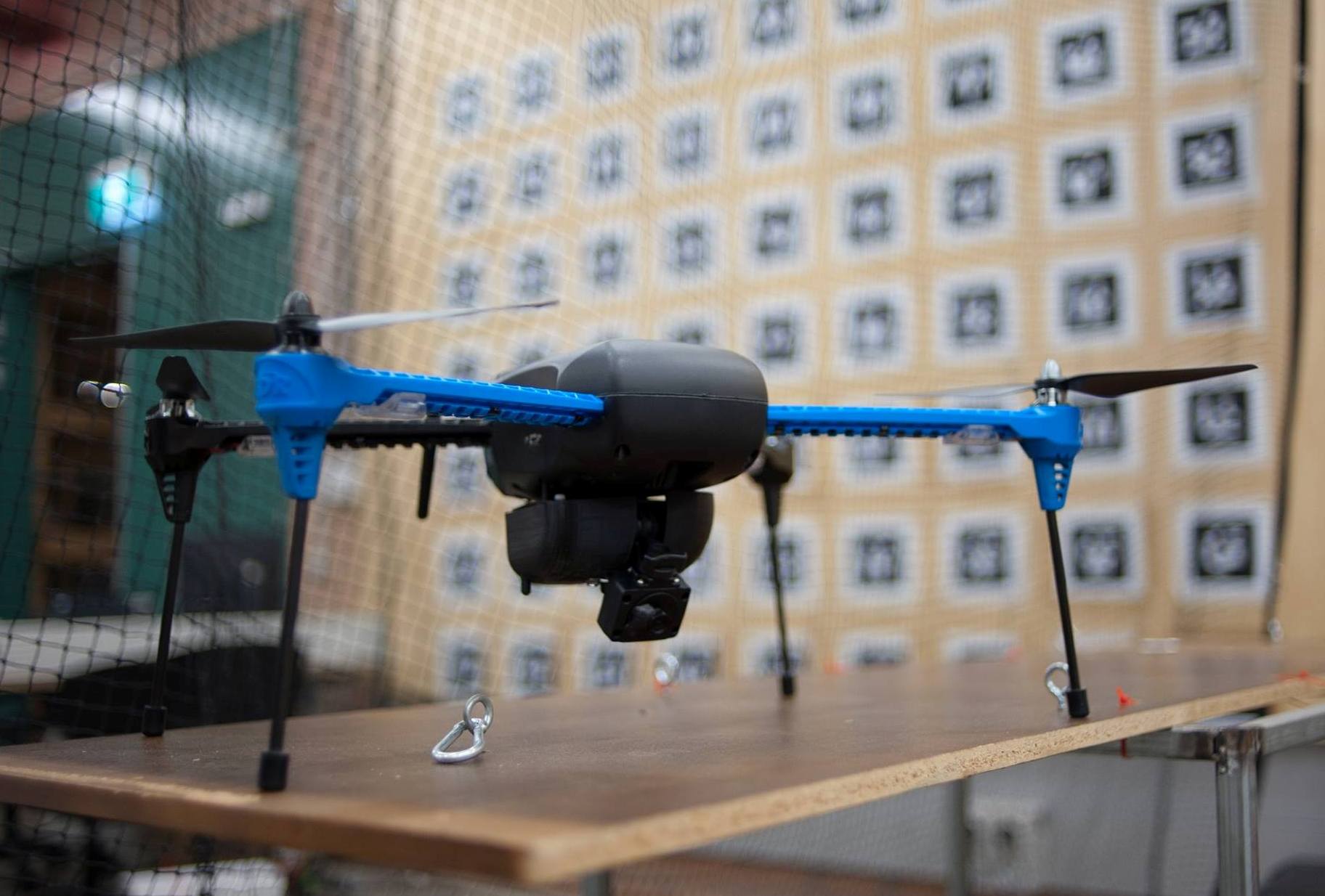}
\caption{}
\label{fig_iris_with_cam}
\end{subfigure}
\begin{subfigure}[b]{0.23\textwidth}
\centering
\includegraphics[width=\textwidth]{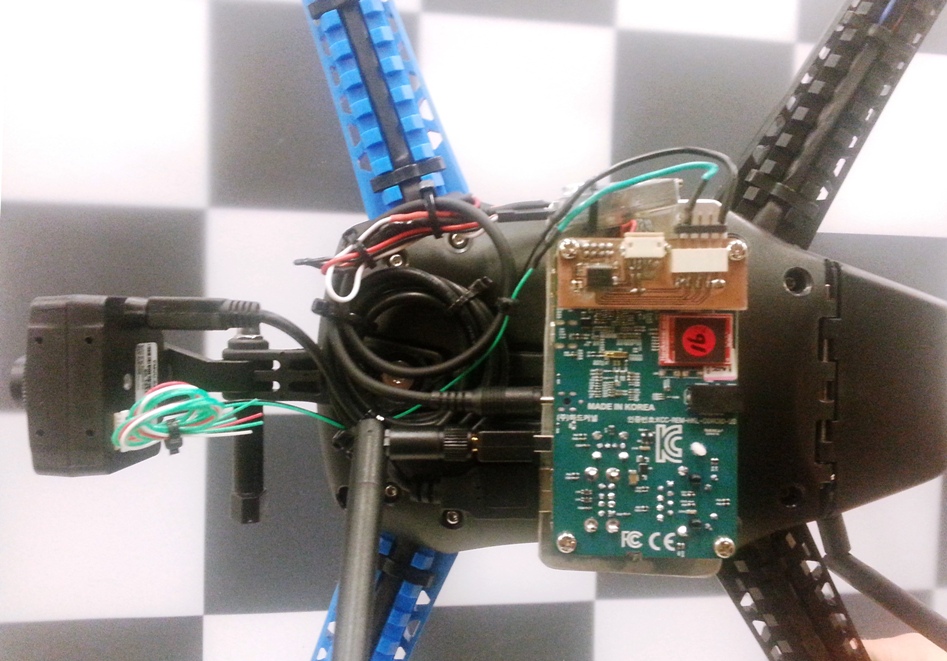}
\caption{}
\label{fig_quad_hardware}
\end{subfigure}
\caption{(\subref{fig_iris_with_cam}) - Iris+ Quadrotor MAV with sensing and computing package for VIO attached underneath. (\subref{fig_quad_hardware}) - Close-up of the sensing and computing hardware illustrating the on-board camera and computer.}
 \label{fig_quad_and_components}
\end{figure}%
Therefore, there exists a need for a low-level estimator for MAVs capable of producing fast and accurate estimates of the platform velocity and attitude. 
To account for this need, our previous work presented a novel estimator design that combines the IMU measurements with the MAV dynamic model \cite{dmw2013}. This design facilitates the drift free estimation of lateral and longitudinal components of quadrotor MAV body frame velocity in addition to its attitude. However, the vertical component of body frame velocity and the accelerometer biases remained unobservable. Addressing this shortcoming is one of the main contributions of this paper.

The main objective of this paper is a visual-inertial odometry system that is capable of producing fast, real-time drift free estimates of the attitude and velocity of a quadrotor MAV, employing monocular images and biased IMU measurements. This system is envisioned as forming a critical component of a low level, fast velocity and attitude feedback control system for the quadrotor MAV. As such the focus here is on an undelayed sensor fusion algorithm with low computational complexity that can operate in real-time on-board an embedded computer with restricted computing resources.

Specifically, this paper makes three main contributions. First, it makes use of pair-wise image feature correspondences to enforce multiple constraints on an EKF that fuses the IMU information with quadrotor MAV dynamic model in a principled way. This was achieved without making a planar environment assumption as was the case in \cite{bristeau2011navigation}. Formulation proposed here enables drift free estimation of velocity and attitude along with all IMU biases, thus improving the utility of the model based state estimators for quadrotor MAVs detailed in \cite{dmw2013}. As our second contribution we identify that the observability of this formulation degrades, irrespective of the information provided by the IMU and the dynamic model, when the quadrotor MAV is stationary, and present suitable modifications to minimise this issue with only a marginal increase in computational complexity. 

Thirdly, via an online implementation on-board a 50 gram embedded computer, we demonstrate that the proposed algorithm is suitable as a fast low-level state estimator for quadrotor MAVs. Exploiting the parallel processing and vector processing instructions of the embedded computer, we demonstrate that the said implementation is capable of processing VGA images as fast as 100Hz while tracking 40 environmental features on average in each image. We also present experimental results obtained through a system level implementation of the proposed algorithm which fuses IMU measurements at 200Hz and camera images at 50Hz in real-time.

\section{Related Work}
This section presents an outline of previous literature in visual-inertial odometry and model-aided state estimation that are most relevant to the work presented here. A more detailed review of fusing visual and inertial measurements can be found in the proceedings of \cite{Dias2007}.

Kinematic and dynamic constraints of the platform to which the inertial sensors are attached have been employed in the literature to aid the state estimation process. Often these are soft constraints, based on the simplified characteristics of the mobile robotic platform under consideration. For example, \cite{dissanayake2001aiding} exploited the nonholonomic constraints governing the motion of a land vehicle moving on a surface to improve the accuracy of roll and pitch estimates obtained by fusing accelerometer and gyroscope measurements. In another example, \cite{koifman1999} illustrated how the aircraft dynamics of a fixed wing MAV can be used to aid a consumer grade INS, so that the estimation accuracy can be improved. 

Martin and Salaun \cite{martin2010} suggested that dynamics of quadrotor MAVs could be used to produce drift free velocity estimates relying only on inertial sensors. They illustrated that in the absence of biases, accelerometer measurements can be low pass filtered to obtain an initial estimate of two components of the quadrotor MAV translational velocity. An attitude and velocity estimator based on this principle was  presented by the authors in \cite{dmw2013}. However, in this design the accelerometer biases were assumed known and only a partial velocity estimate was available. 

Using information from a monocular camera can significantly improve the quadrotor MAV pose estimation results. There are two main algorithms to achieve this without having to estimate the environmental structure. One approach is to find environmental feature observations common to a pair of images, derive the fundamental matrix that describe the relative motion between the two camera poses and then decompose that to obtain a relative pose estimate. Williams and Reid \cite{williams2010combining} argued that this method has several disadvantages when the requirement is to design a real-time state estimation algorithm on a computationally constrained platform. Instead, they made use of individual updates from each corresponding point pair. These updates were performed in an EKF, by making use of epipolar geometry between two consecutive images to define constraints on the relative camera motion based on each pair of point correspondences. They demonstrated that considerable improvements in estimation accuracy can be achieved in a Visual SLAM algorithm by incorporating multiple such constraints. However, they did not employ IMU measurements which add additional complexity to the filter design. 

Similarly in \cite{webb2007vision}, epipolar geometry based constraints are combined with several simplified dynamic models that describe the motion of a fixed-wing MAV in an iterated EKF. There, simulations were employed to demonstrate that incorporating a more accurate dynamic model results in considerable improvements in filter accuracy.  However, as will be shown later, relying on visual constraints between the current frame and the immediate previous frame results in observability issues when the camera is stationary. This can severely hamper the state estimation accuracy of quadrotor MAVs which tend to spend a considerable amount of flight time in or near hover. In Section \ref{sec_key_frames}, we present a detailed analysis of this issue and a computationally efficient method to overcome it.

Several examples of combining visual constraints induced by corresponding point pairs with inertial measurements can also be found in the literature. In \cite{diel2005epipolar}, measurements from a gyroscope triad are employed to remove the relative rotation between two consecutive images. The corresponding point pairs between the two warped images are then employed to define constraints between the camera positions. These constraints are incorporated as measurements into an EKF which also fuses accelerometer measurements. This approach may result in slow drift in orientation estimates as the gyroscope biases are neglected in the image warping process. Further, they also do not address the observability issues that occur when the camera is stationary. In \cite{qayyum2012seamless}, feature point correspondences between images are employed to derive directional constraints that are then incorporated as additional measurements to improve the estimation accuracy of a Visual SLAM algorithm. However, they neglected information available in the point correspondences about camera rotation, which may result in loss of accuracy. 

In \cite{mourikis2007multi} an EKF design that incorporates visual and inertial information without modelling the environmental structure is presented. This was achieved by augmenting the state vector with $N$ previous camera poses, so that all environmental features that are observed from those poses can be employed to define constraints among the $N$ poses. This approach, known as MSCKF, was further improved in \cite{li2013high} and \cite{hesch2014} such that appropriate observability constraints are enforced, resulting in improved estimation accuracy. Though the environmental features are not included in the state vector in the different versions of MSCKF, that design still relies on a least-squares estimation of 3D feature position, resulting in additional computational complexity. Further, the information from the visual measurements are incorporated into the filter in a delayed manner, after each feature has been observed by multiple camera poses. Such delayed fusion of measurements is not suitable for MAVs such as quadrotors with fast dynamics operating in close proximity to obstacles. The EKF algorithm proposed in this paper performs undelayed measurement updates, and also applies the ``First Estimates Jabobian EKF" as in \cite{li2013high} when performing updates using image features. 

In \cite{faessler2015autonomous}, the authors made use of a method known as semidirect visual odometry (SVO) to generate fast (70Hz) estimates of the pose of a quadrotor MAV pose using an on-board camera and an embedded computer. However, these pose estimates are only accurate up-to-a-scale. Therefore, the authors then combined these estimates with inertial measurements in a separate EKF. The results presented therein demonstrate that this method is capable of producing both fast and accurate state estimates. However, this loosely-coupled combination of the scale-less pose estimates and IMU measurements require careful manual initialization and filter turning, as the uncertainty of SVO pose estimates in metric scale is unknown prior to filter initialization. Though this initialization can be performed once prior to take-off, any subsequent failure in tracking due to fast camera motion necessitates a re-initialization which cannot be performed on the fly once the quadrotor is airborne. In contrast, the state estimator proposed in this paper does not require such initialization as the visual measurements are directly made on the images where their uncertainty can be known beforehand.

Closer in spirit to the work presented here is the work by Bristeau et al.\ in which they combined the quadrotor dynamic model with visual and inertial measurements to design a state estimator for the AR Drone quadrotor MAV \cite{bristeau2011navigation}. However, in formulating the estimator, they made use of two restrictive assumptions about the environment in which the quadrotor MAV is operating. First, they assumed that all environmental features lie on a plane and derived a homography matrix from point correspondences between image pairs. Second, they assumed that the distance from the MAV to the said plane is known and employed that information to derive metric scale velocity estimates. Neither of these assumptions are employed in the work presented here and as such the estimator proposed here can be deployed in a wider variety of real-world environments and flight envelopes.

\section{Estimator Design}
\subsection{Frame Definition and State Vector}
The system under consideration is a quadrotor MAV affixed with an IMU (consisting of a triad of orthogonal accelerometers and gyroscopes) and a monocular camera. We define a body fixed coordinate frame $\{B\}$  with origin at the centre of mass of the quadrotor and aligned such that the $^B\bm x,\,^B\bm y$ plane is parallel to the propeller plane. We also assume that the IMU is located at the centre of gravity of the quadrotor and aligned with $\{B\}$. Camera coordinate frame is defined as $\{C\}$ and for clarity we assume that its origin coincide with that of $\{B\}$. Also defined is an earth fixed inertial coordinate system $\{W\}$, which is defined by the position and orientation of $\{B\}$ at the starting position of the estimator. $^B_WR$ denotes the rotation of $\{B\}$ with respect to $\{W\}$ and $^C_BR$ is the rotation of $\{C\}$ with respect to $\{B\}$ which is assumed known and fixed. 

Vectors are denoted in boldface. A trailing subscript denotes what is being measured. A leading subscript denotes frame in which a vector is being expressed. Also, a second trailing subscript is used to denote specific elements of a vector. For example $v_{Bx}$ denotes the first component of $\bm v_B$.

The state $\bm x$ is defined as:
\begin{align*}
\bm x = \begin{bmatrix}
\,^W\bm p_B & \bm \Theta & \bm v_B & \bm\beta_a & \bm\beta_g
\end{bmatrix}^T.
\end{align*}
where $^W\bm p_B$ is the position of the origin of $\{B\}$ in $\{W\}$, $\bm \Theta = [\phi\,\,\theta\,\,\psi]^T$ is the orientation of $\{B\}$ with respect to $\{W\}$ in ZYX Euler angle parametrization, $\bm v_B$ is the velocity vector of the origin of $\{B\}$ measured with respect to $\{W\}$ and expressed in $\{B\}$, $\bm\beta_a$ is the accelerometer bias and $\bm\beta_g$ is the gyroscope bias. Given the real-time requirements of the state estimator, we propose an EKF to estimate the state $\bm x$ by fusing the available measurements. The following section details the process and measurement equations for the EKF.

\subsection{Process Model}
A detailed description of the dynamics of the quadrotor MAV can be found in \cite{mahony2013}. A simplified model of the translational dynamics in a form suitable for state estimation was presented in \cite{dmw2013}. The main characteristic of this dynamic model is the presence of a drag force that is approximately proportional to the projection of MAV body frame velocity $\bm v_B$ on to the propeller plane. Here, we present a slightly more accurate form and refer interested reader to \cite{mahony2013} and \cite{dmw2013} for further information. 

The process equation governing the evolution of the state vector $\bm x$ is given by:
\begin{align}\label{eq_proc_1}
\begin{bmatrix}
\,^W\dot{\bm p}_B \\
 \dot{\bm \Theta} \\
 \dot{\bm v}_B \\
 \dot{\bm\beta}_a\\
 \dot{\bm\beta}_g
\end{bmatrix}
&=
\begin{bmatrix}
\,^B_WR\bm v_B \\
\Xi(\bm \omega_g - \bm \beta_g + \bm \eta_g) \\
\,^W_BRg\bm e_3 - \bar{D}_L \bm v_B  + \bm f_{ip}+ \bm\eta_v \\ 
\bm 0\\
\bm 0
\end{bmatrix}
\end{align}
where $ \bm f_{ip} = (\,^Ba_z - \beta_{az})\bm e_3 - (\bm \omega_g - \bm \beta_g + \bm \eta_g)\times \bm v_B$,
$g$ is the magnitude of gravity, $\bm e_3$ is a unit vector with a 1 in the third element, $\bar D_L = k_1 (\bm I_3 - \bm e_3^T \bm e_3)$ with $k_1$ a constant and $\bm I_3$ the $3\times 3$ identity matrix, $m$ is the mass of the quadrotor and $\bm\eta_v$ is a White Gaussian Noise (WGN) vectors denoting the errors in the quadrotor MAV dynamic model. Also $\Xi$ denotes the matrix that relates body rotational rate to the Euler rates.

Eq. (\ref{eq_proc_1}) makes use of the fact that the thrust force $f_T$ and the rotational velocity ${\bm \Omega}_B$ can be replaced by the $^B\bm{z}$ axis accelerometer measurement $\,^Ba_z$ and gyroscope measurements $\bm\omega_g$, respectively as:
\begin{align*}
{\bm \Omega}_B = \bm \omega_g - \bm \beta_g + \bm \eta_g
&, \,\,\,\,\,\, \frac{f_T}{m} = \,^Ba_z - \beta_{az} 
\end{align*}
where $\bm \eta_g$ is the measurement noise assumed to be WGN \cite{dmw2013fsr}. 

\subsection{Measurement Model}
Measurement of the $^B\bm x, ^B\bm y$ accelerometers are derived from the quadrotor dynamic model:
\begin{align*}
\bm h_{a} &= \Upsilon(- \bar{D}_L \bm v_B + {\bm\beta}_a + \bm \eta_a )
\end{align*}
where $\Upsilon$ is a matrix that extracts the first two elements of a $3\times 1$ vector and $\bm \eta_a$ is a $3\times 1$ vector of WGN denoting the accelerometer measurement noise \cite{dmw2013}.

The information from the camera images are incorporated into the filter using the epipolar geometry constraint between the current image (taken at time $t_c$) and the previous image (taken at time $t_p$). This method, as opposed to the batch update methods that rely on the decomposition of the fundamental matrix, enables us to perform visual measurement updates irrespective of the number of corresponding point pairs between the two images. Additionally, this facilitates a filter design where visual measurement updates can be performed with as many point pairs as possible until the measurements from the next image becomes available. This results in a streamlined filter implementation that is able to exploit the benefits of multi-core processor architectures. More details on such an implementation can be found in Section \ref{sec_exp}.

The epipolar constraint between a pair of images taken at times $t_c$ and $t_p$ can be expressed as:
\begin{align}
0 &= (\bm p_i^{t_c})^TK^{-T}EK^{-1}\bm p_i^{t_p} \nonumber \\
&= (\bm p_i^{t_c})^TK^{-T}\left[\bar{\bm p}\right]_\times \bar{R}K^{-1}\bm p_i^{t_p}\label{eq_epipolar_true} 
\end{align}
where $\bar{\bm p} = \,^W_CR(t_c)(\,^W\bm p_C(t_p) - \,^W\bm p_C(t_c))$ and $\bar{R} =\,^W_CR(t_c)\,^C_WR(t_p)$ and $K$ is the intrinsic camera calibration matrix. $R(t_c)$ denotes the rotation matrix constructed using the orientation at time $t_c$. Also $\bm p_i^{t_c}$, and $\bm p_i^{t_p}$ are homogeneous coordinates of corresponding feature points between the current and the previous images, respectively.  $[\bm p]_\times$ creates the skew symmetric matrix of $\bm p$ such that $[\bm p]_\times\bm q = \bm p \times \bm q$ for any given $3\times 1$ vectors $\bm p, \bm q$. The expression for the essential matrix $E$ can be further simplified by defining $\bm p_d = \,^W\bm p_C(t_p) - \,^W\bm p_C(t_c)$. Then:
\begin{align*}
E &=\left[\bar{\bm p}\right]_\times \bar{R}
= \,^W_CR(t_c)\left[\bm p_d\right]_{\times}\,^C_WR(t_p).
\end{align*}

Within the EKF, the epipolar constraint is constructed using the current best estimate of the state vector $\hat{\bm x}$. The errors in both the state estimate and the feature point tracking results in a deviation from the result predicted in Eq. (\ref{eq_epipolar_true}):
\begin{align*}
\hat{h}_v &= (\bm p_i^{t_c})^TK^{-T}\hat{E}K^{-1}\bm p_i^{t_p}
\end{align*}
which is then used in the update step of the EKF. Here, $\hat{E}$ is the essential matrix constructed using the estimated state.

\subsection{State Augmentation}
For the visual measurement update, the pose of the camera where the previous image was captured is required. Since this is not in the state vector discussed above, the state needs to be augmented with this pose at the end of incorporating measurements from each image. The new state vector is then given by: $\bm x = \begin{bmatrix}
\,^W\bm p_B & \bm \Theta & \bm v_B & \bm\beta_a & \bm\beta_g & ^W \acute{\bm p}_B & \acute{\bm \Theta} 
\end{bmatrix}^T$.
Since the previous pose has zero dynamics, the new process equation can be obtained by augmenting Eq. (\ref{eq_proc_1}) with $ ^W\dot{ \acute{\bm p}}_B = 0$ and $ \dot{\acute{\bm \Theta}}= 0$. 

Given the new states, the visual  measurement equation can be re-written as:
\begin{align}\label{eq_vision_measurement}
\hat{h}_v &= (\bm p_i)^TK^{-T}\hat{E}K^{-1}\acute{\bm p}_i
\end{align}
where $\hat{E} =\,^B_CR\,^W_BR( \hat{\bm \Theta})\left[\bm \hat{\bm p_d}\right]_{\times}\,^B_WR(\hat{\acute{\bm \Theta}})\,^C_BR$
and $ \hat{\bm p}_d = \,^W \hat{\acute{\bm p}}_B - \,^W \hat{\bm p}_B$. Also $R(\hat{\bm \Theta})$ denotes a rotation matrix constructed from $\bm \hat{\bm \Theta}$. 

As both $\bm p_i$ and $\acute{\bm p}_i$ are random variables, ideally those should be incorporated into the filter state prior to performing the updates from each point pair, and then marginalized after the update is performed. However, Soatto et al.\ \cite{soatto1996motion} demonstrated that this additional computational complexity results in only marginal improvement in filter performance. Therefore, we make use of the approach used in \cite{williams2010combining} and \cite{webb2007vision} to derive a model for the noise due to errors in $\bm p_i$ and $\acute{\bm p}_i$. For this purpose, Eq. (\ref{eq_vision_measurement}) can be re-written as:
\begin{align*}
\hat{h}_v &= (\bm p_i)^T\bm g(\hat{\bm x}, \acute{\bm p}_i)
\end{align*}
where $\bm g(\hat{\bm x}, \acute{\bm p}_i) =K^{-T}\hat{E}K^{-1}\acute{\bm p}_i$. Assuming that the noise in the image measurement $\bm p_i$ is represented by the covariance matrix $R_{im}$, the measurement covariance can be computed as:
\begin{align*}
R_{vo} &= \frac{\partial h_v}{\partial p_i}R_{im}(\frac{\partial h_v}{\partial p_i})^T
= \bm g^T(\hat{\bm x}, \acute{\bm p}_i)R_{im}\bm g(\hat{\bm x}, \acute{\bm p}_i).
\end{align*}
Similar to the approach in \cite{williams2010combining}, we assume that there is no uncertainty in the measurement $\bm \acute{\bm p_i}$ as the position of the feature in the previous image is known exactly. Also assuming that the standard deviation for image measurements on the current image is $\sigma_i$, $R_{im}$ can be expressed as: $R_{im} = diag(\begin{bmatrix}\sigma^2_i & \sigma^2_i & 0\end{bmatrix})$.


\subsection{EKF Mechanization Equations}
EKF prediction and accelerometer measurement updates follow the standard formulation \cite{grewal2001}. The visual measurement update is performed for each corresponding image point pair. The epipolar constraint is incorporated into the EKF update using the perfect measurement method \cite{simon2010kalman}. Outliers in feature point correspondences were identified and removed during visual measurement update by applying a threshold of $2\sigma_i$ on the magnitude of innovation for each corresponding point pair. Details are omitted here for brevity.

\subsection{Key-frame Based Updates}\label{sec_key_frames}
As will be demonstrated in Section \ref{sec_simulation}, with the above formulation of the filter, the states that are not constrained by the accelerometer updates begins to diverge when the vehicle is stationary. The reason for this behaviour can be explained using Eq. (\ref{eq_vision_measurement}). When the vehicle (or camera) is stationary, the pixel locations of corresponding points  $\bm p_i$ and $\acute{\bm p}_i$ are equal up to a small tracking error. When $\bm p_i \approx \acute{\bm p}_i$, then $\hat{h}_v \approx 0$ for any given essential matrix E. For this reason, any errors in the predicted essential matrix $\hat{E}$ due to the errors in the state vector are not reflected in the visual measurement $\hat{h}_v$. Under such a situation, the visual measurements do not contain any information about the state vector $\bm x$ and therefore, those states that solely rely on visual measurement updates will diverge despite repeated updates. As illustrated in Fig. \ref{velocity_error} in Section \ref{sec_simulation}, this deviation is most prominent in the z axis position estimate, as that axis, unlike the x and y axes, is not complemented by the velocity measurements obtained through the accelerometer and the MAV dynamic model.


The filter formulation proposed here is augmented with the current pose of the camera after performing all updates corresponding to a given image. All future visual measurements will be made with respect to this augmented state. For ease of reference, the images for which the filter state is augmented are termed key-frames, borrowing the concept of key-frames from previous literature \cite{nutzi2011}. We make two improvements to how and when key-frames are initialized to minimize the divergence due to a stationary camera. First, key-frames are initialized using a direct feature disparity calculation between the previous key-frame and the current image, thus ensuring sufficient disparity in most situations. Second we perform state augmentation (including visual measurement update) only when a new key-frame is initialized. For all other images obtained prior to the next key-frame initialization, visual measurement update is still performed, albeit without state augmentation.

There are multiple ways to calculate the feature disparity $f_d$ between two corresponding point sets. We opted for a simple approach of calculating the mean of the absolute value of the disparity vector according to $f_d = \frac{1}{n_f}\Sigma_{i=1}^{n_f}\|\bm p_i - \acute{\bm p}_i\|$ where $n_f$ is the number of corresponding points and $\bm p_i$, $\acute{\bm p}_i$ are corresponding elements of the point sets. Although this proved to be adequate, arguably, better forms for calculating the disparity could exist. 
%
%

\section{Simulation Results}\label{sec_simulation}
This section presents the estimation results for the data obtained from a simulation run of the quadrotor MAV simulator, modelled according to the dynamics presented in \cite{mahony2013}. The purpose of the simulation results presented here is to demonstrate consistency and the observability properties of the proposed estimator. 

The simulator operates the quadrotor along a given trajectory and records accelerometer and gyroscope measurements at 200Hz assuming $\eta_{ax} \in \mathcal{N}(0,0.25 m^2s^{-4})$ and $\eta_{gx} \in \mathcal{N}(0,0.005 \textup{rad}^2s^{-2})$ and similar values for the remaining components of accelerometer and gyroscope noise. The accelerometer and gyroscope measurements were also corrupted by random constant biases. Throughout the trajectory, images were generated at 10Hz from a pinhole camera assumed to be on-board the quadrotor MAV, with its optical axis aligned with $^B\bm x$. For the purpose of creating the images, a virtual world size $200\times 200\times 50m$ consisting of 2000 uniformly distributed point features was created in the vicinity of the MAV. 

The state estimator was implemented in Matlab assuming perfect knowledge of sensor noise covariances and camera calibration matrix used to generate the images. FAST features \cite{rosten2006} are tracked between consecutive images using the Matlab implementation of Kanade-Lucas-Tomasi (KLT) feature tracker \cite{tomasi1991}. If the number of features tracked falls below 30, then new FAST features are detected in the current image and added to the tracker. Also a suitable value for the tracking error $\sigma^2_i$ was obtained through experimentation.

\begin{figure}[htb]
\begin{subfigure}{0.23\textwidth}
\centering
\includegraphics[trim = 0cm 1.2cm 0cm 2.4cm, clip, width=1\textwidth]{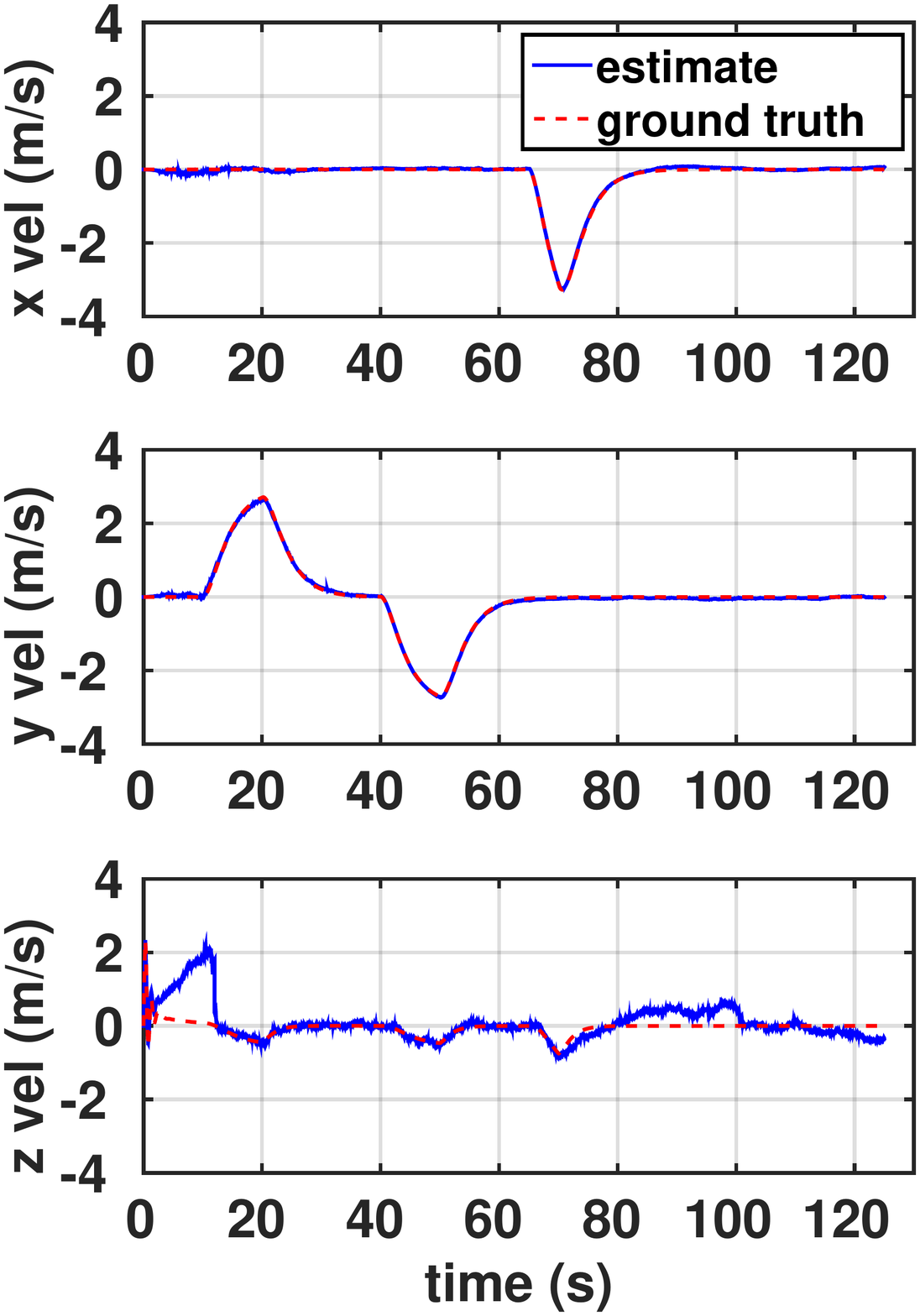}
\caption{Velocity}
\label{velocity}
\end{subfigure}
\begin{subfigure}{0.23\textwidth}
\centering
\includegraphics[trim = 0cm 1.2cm 0cm 2.4cm, clip, width=1\textwidth]{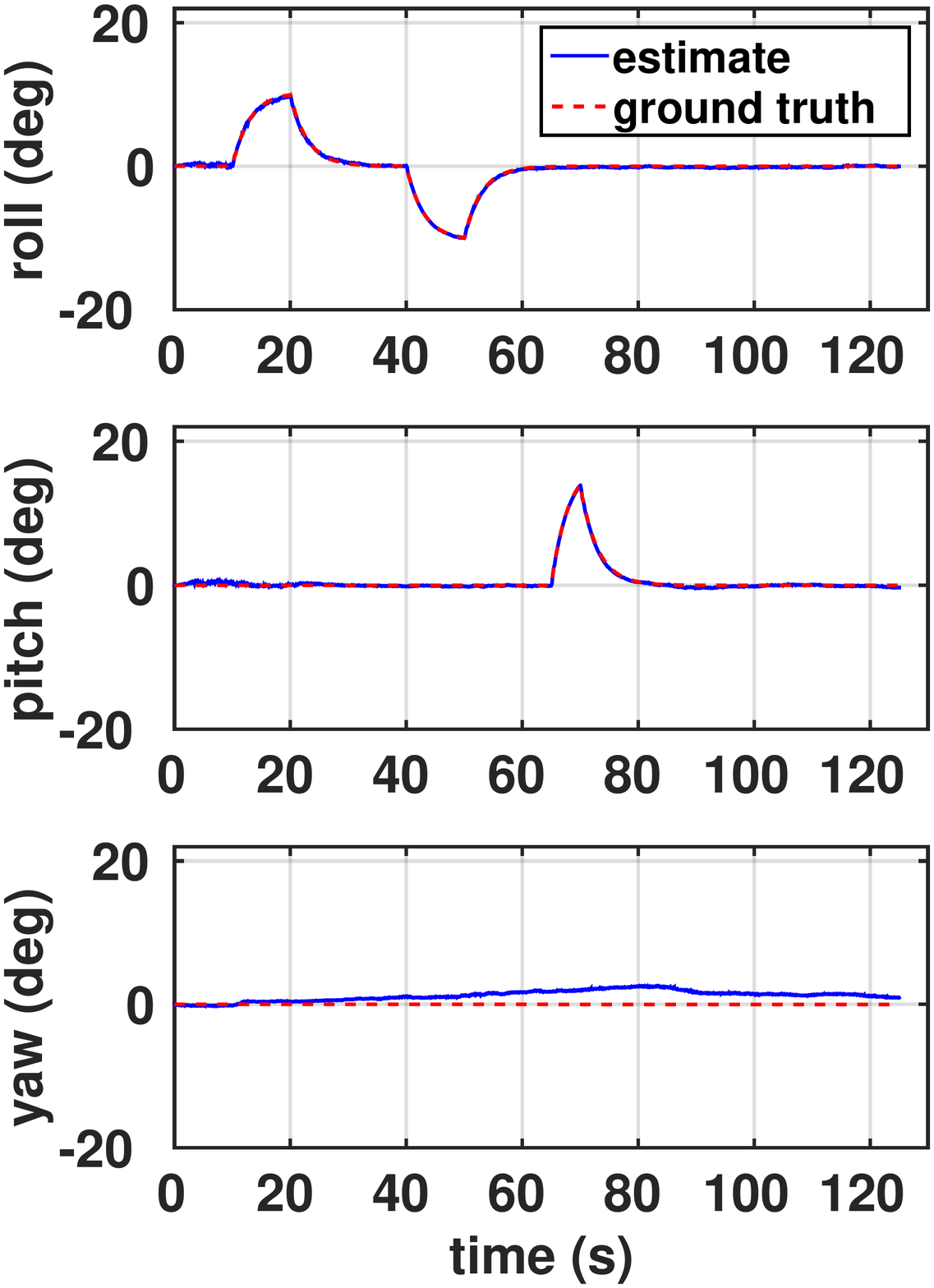}
\caption{Orientation}
\label{angle}
\end{subfigure}
\caption{True and estimated velocity and orientation for simulations}
 \label{fig_pos_vel}
\end{figure}
\begin{figure}[tb]
\centering
\begin{subfigure}[b]{0.23\textwidth}
\includegraphics[trim = 0cm 1.2cm 0cm 2.4cm, clip, width=\textwidth]{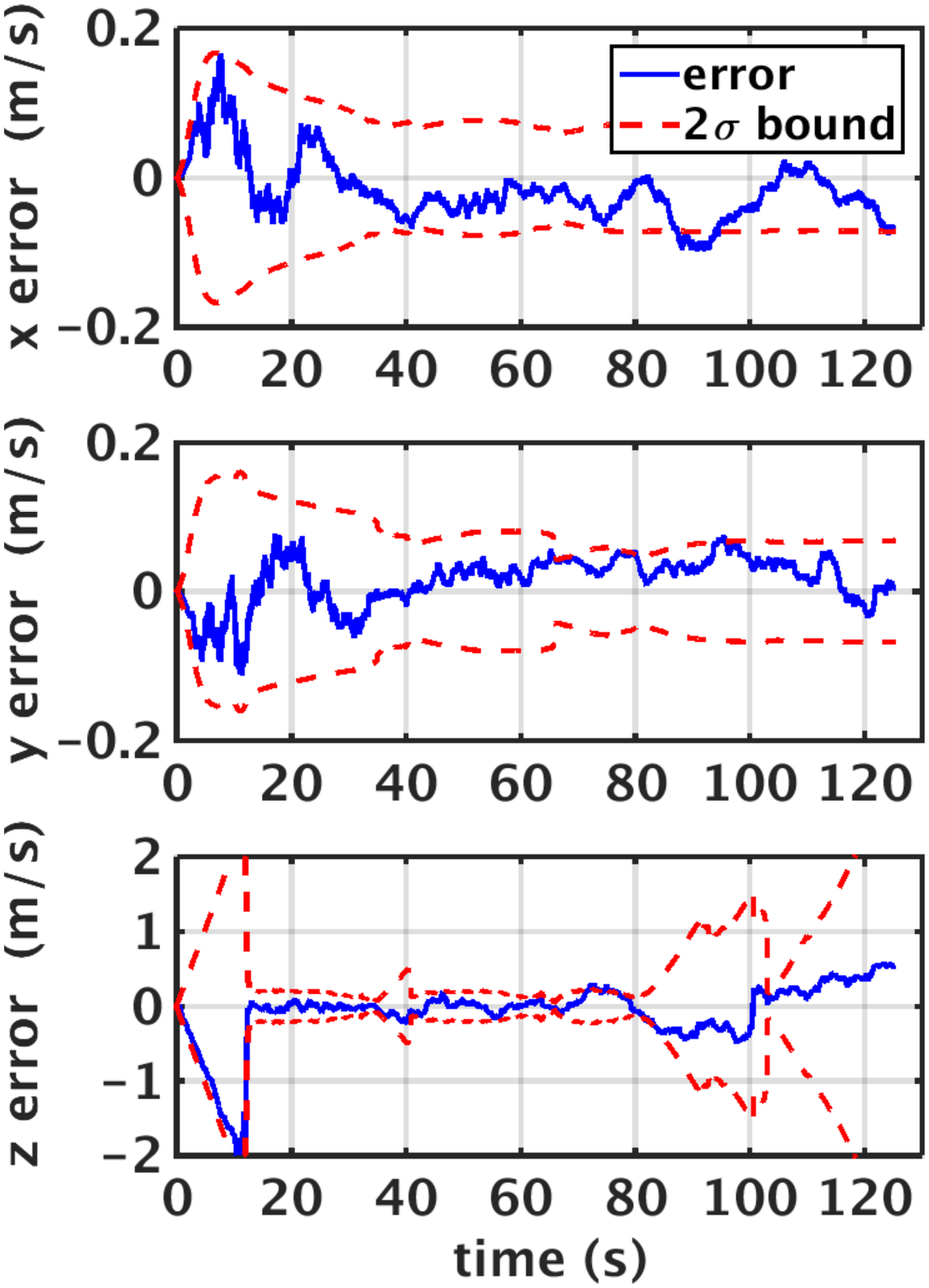}
\caption{With vision updates}
\label{velocity_error}
\end{subfigure}
\begin{subfigure}[b]{0.23\textwidth}
\includegraphics[trim = 0cm 1.2cm 0cm 2.4cm, clip, width=\textwidth]{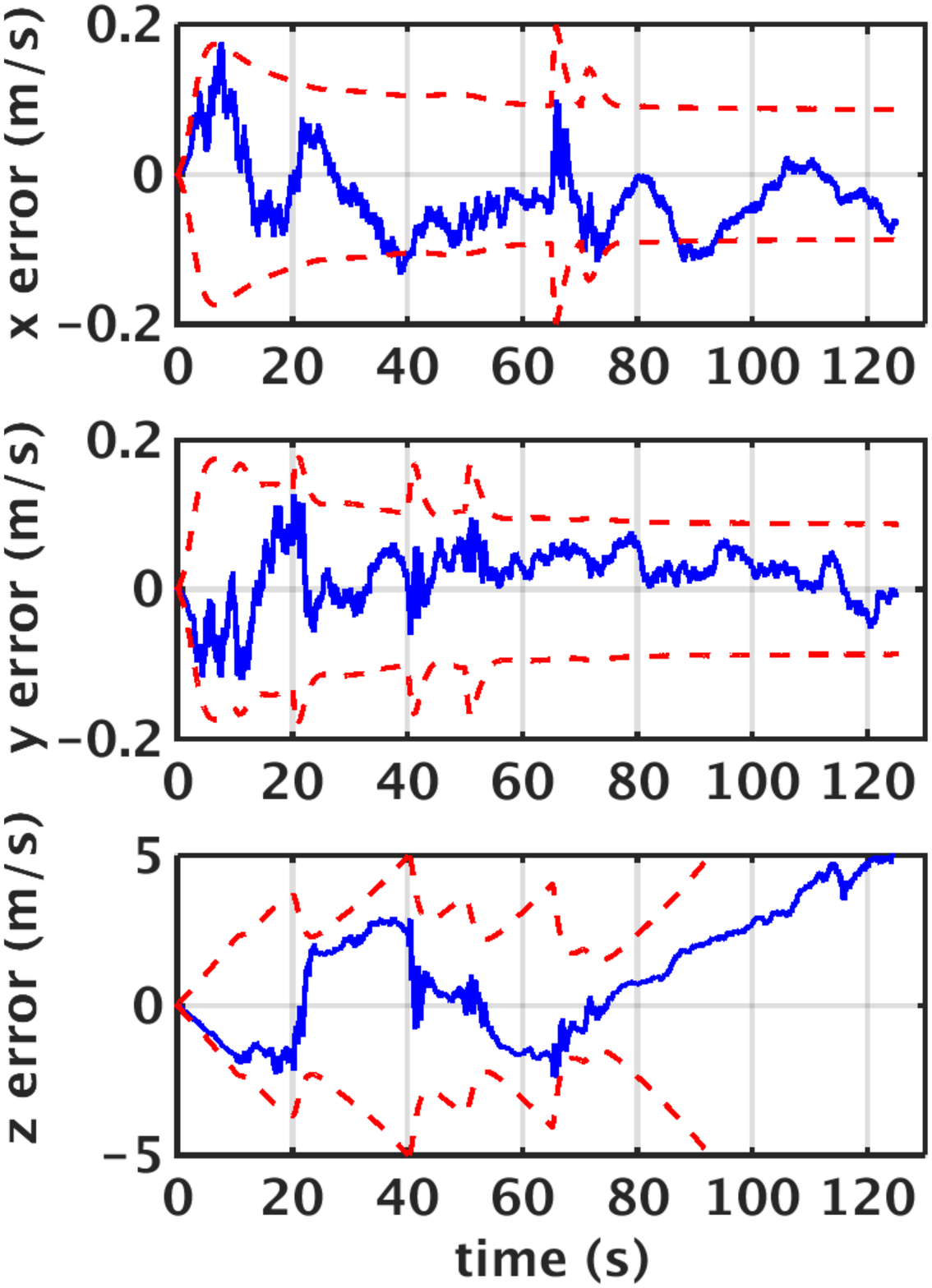}
\caption{Only inertial updates}
\label{velocity_error_iner_only}
\end{subfigure}
\caption{Velocity estimation errors with $2\sigma$ bounds for simulations without key-frames. Note the scale difference in z axis.}
 \label{fig_vel_err}
\end{figure}

Fig. \ref{fig_pos_vel} illustrates the estimated velocity and orientation along with the ground truth. Fig. \ref{fig_vel_err} illustrates the velocity  estimation errors with $2\sigma$ bounds. For comparison purposes, it includes the estimation errors obtained from the same estimator if vision updates were disabled. Comparing Fig. \ref{velocity_error} with Fig. \ref{velocity_error_iner_only}, it can be seen that the accuracy of velocity estimate improve considerably, when the visual measurement updates are employed. This is most prominent along the $^B\bm z$ axis, for which the velocity estimates are not aided by the quadrotor dynamic model. 

\begin{figure}[htb]
\centering
\begin{subfigure}[t]{0.23\textwidth}
\includegraphics[trim = 0cm 1.2cm 0cm 2.4cm, clip, width=1\textwidth]{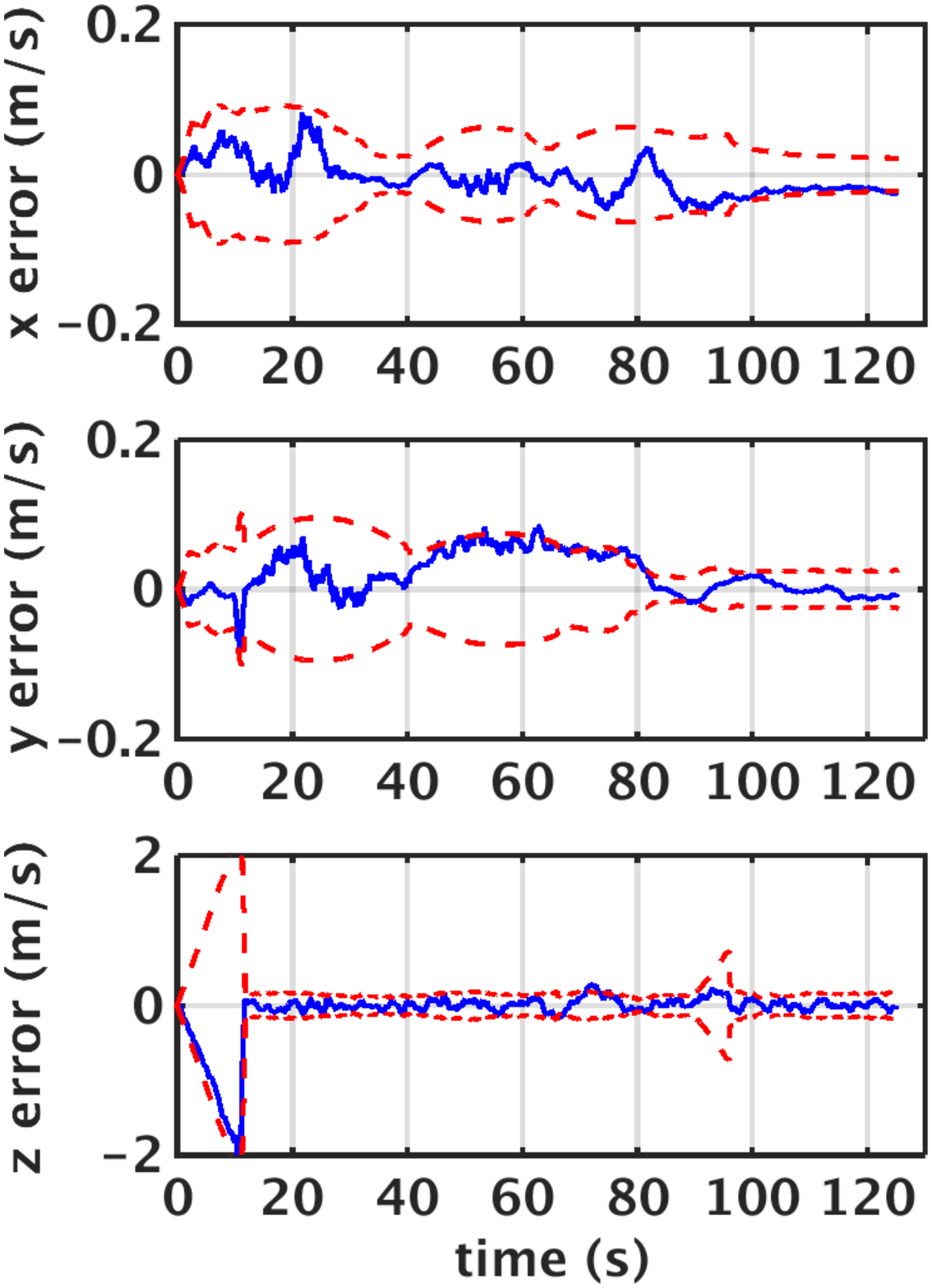}
\caption{Velocity estimation error}
\label{vel_error_keyframe}
\end{subfigure}
\centering
\begin{subfigure}[t]{0.23\textwidth}
\centering
\includegraphics[trim = 0cm 1.2cm 0cm 2.4cm, clip, width=1\textwidth]{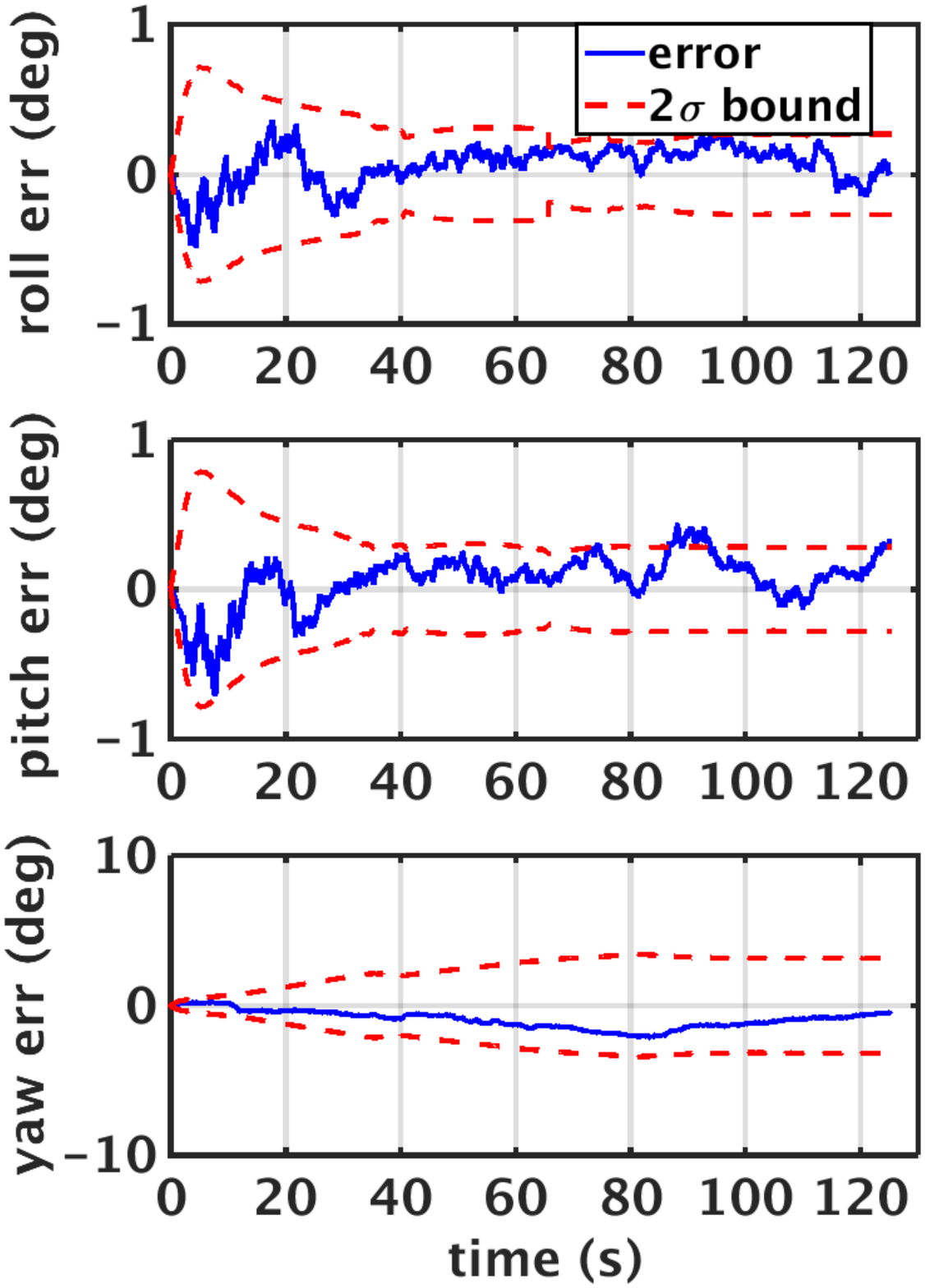}
\caption{Orientation estimation error}
\label{fig_ang_err}
\end{subfigure} %
\caption{Velocity and orientation estimation errors and their respective $2\sigma$ bounds with key-frames. Note the scale difference in z axis.}
 \label{fig_pos_err_keyframe}
\end{figure}
Fig. \ref{fig_vel_err} also illustrates how the $^Bv_z$ estimate begins to drift when the MAV becomes stationary approximately 80 seconds after take-off. Fig. \ref{vel_error_keyframe} demonstrate how this drift in velocity estimates can be overcome by the key-frame based update approach detailed in Section \ref{sec_key_frames}. A direct comparison of Fig. \ref{velocity_error_iner_only} and Fig. \ref{vel_error_keyframe} reveals the full extent of the improvement in velocity estimation accuracy brought about by the proposed state estimator design. 

Fig. \ref{fig_ang_err} demonstrates the errors and $2\sigma$ bounds for the orientation estimates of the proposed filter employing key-frames. The roll and pitch angle estimation errors remain bounded throughout the flight but the yaw angle estimation error increases while the quadrotor is in motion as both the visual and inertial measurements only provide information about the relative change in yaw. However, when the quadrotor becomes stationary at around 80 seconds, the uncertainty in yaw angle estimate ceases to increase and the errors remain bounded while the quadrotor is in hover, due to the use of key-frames.

\begin{figure}[htb]
\centering
\begin{subfigure}[b]{0.23\textwidth}
\includegraphics[trim = 0cm 1.2cm 0cm 2.4cm, clip, width=\textwidth]{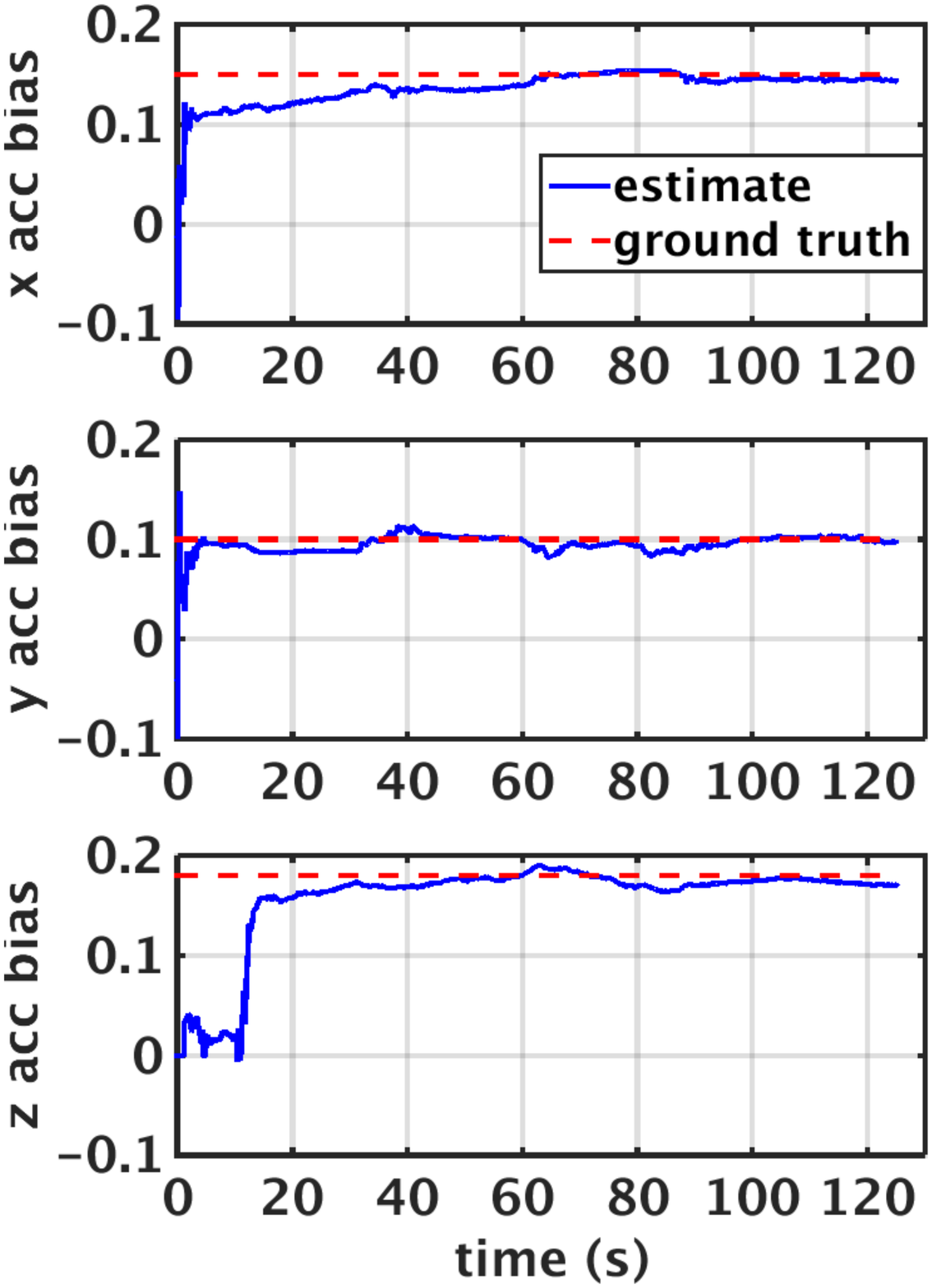}
\caption{Accelerometer biases}
\label{fig_acc_bias_sim}
\end{subfigure}
\begin{subfigure}[b]{0.23\textwidth}
\includegraphics[trim = 0cm 1.2cm 0cm 2.4cm, clip, width=\textwidth]{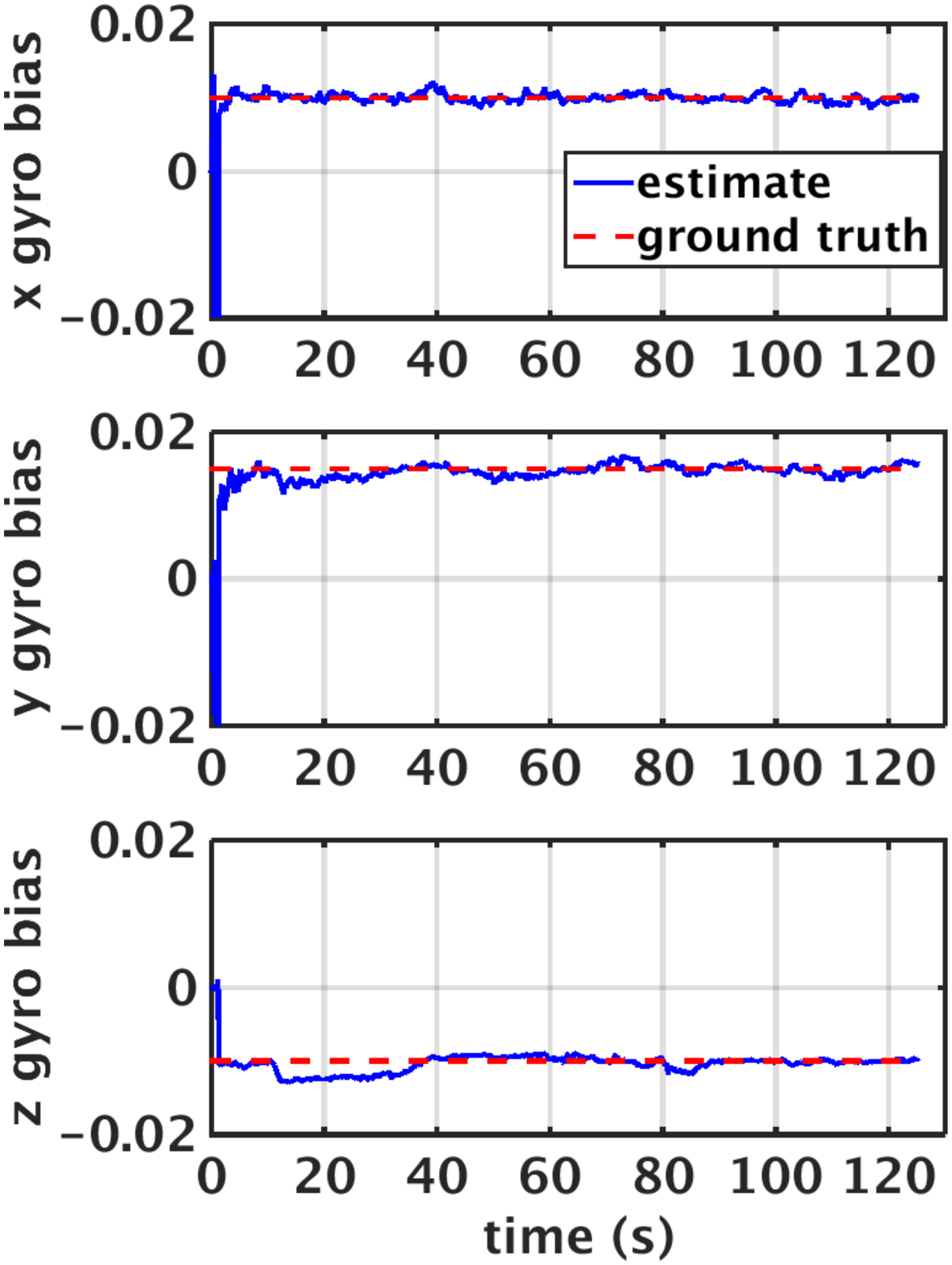}
\caption{Gyroscope biases}
\label{fig_gyro_bias_sim}
\end{subfigure}
\caption{IMU bias estimates with vision updates along with ground truth for simulations}
 \label{fig_bias_sim}
\end{figure}

The bias estimates presented in Fig. \ref{fig_bias_sim} indicate that the IMU biases can be estimated accurately with the proposed estimator formulation. More interestingly $^B\bm z$ accelerometer bias estimates (Fig. \ref{fig_acc_bias_sim}) and velocity estimation errors (Fig. \ref{vel_error_keyframe}) reveal a little known phenomena about state estimators that employ the epipolar constraint. For the first 10 seconds, the simulated quadrotor MAV only exhibits a pure vertical motion (along $^B\bm z$) during which the $^Bv_z$ and $\beta_{az}$ estimates fail to converge. This can be explained in a manner similar to the explanation in Section \ref{sec_key_frames}. During pure vertical motion, any errors in the $^B\bm z$ position estimate of the camera results in the corresponding feature points in the current image moving parallel to their corresponding epipolar lines. As the visual measurement to the EKF is the perpendicular distance of the feature point to its corresponding epipolar line, these parallel displacements do not contribute any information to constrain the errors in $^B\bm z$\footnote{In general, this phenomena is not restricted to pure vertical motion. Motion purely along one direction will result in a divergence of the velocity along that direction. However, this divergence would be much less prominent along $^B\bm x$ and $^B\bm y$ axes as the velocity estimates along them are aided by the quadrotor MAV dynamic model.}.

The $^B\bm z$ velocity estimate only converges when the quadrotor MAV begins to translate along the $^B\bm y$ at 10 seconds, at which point errors in the $^B\bm z$ position estimate causes feature points to be displaced perpendicular from their corresponding epipolar lines. For the same reason, the bias of $^B\bm z$ axis accelerometer only converges after the first 10 seconds has elapsed. In practise however, such situations rarely occur as estimation and control errors will not, in general, allow for motion purely along a single direction, even during take-off (for example, see Fig. \ref{fig_3d_path}).

\section{Experiments}\label{sec_exp}
Flight experiments to evaluate the proposed state estimator were conducted using the Iris+ quadrotor by 3D Robotics Inc. The Iris+ features the open-source Pixhawk auto-pilot containing the Invensense MPU 6000 IMU. An Odroid U3 by Hardkernel was employed as the on-board computer for the Iris+. It features a 1.7GHz Exynos4412 Prime Cortex-A9 Quad-core processor with 2Gbyte main memory. The U3 weighs 48g including heat sink making it an ideal on-board computer for light-weight MAVs. The on-board camera was a Point Grey Firefly which connected to the U3 via a USB interface (see Fig. \ref{fig_quad_and_components}). This camera supports both global shutter and external triggering, eliminating the need for explicit time synchronization between the camera and IMU. The camera was mounted on the Iris+ facing forward, such that its optical axis aligned with  $^B\bm x$.

The proposed state estimator was implemented as a multi-threaded C++ package on the U3, making use of ROS middleware for communications with the camera and the Pixhawk. The feature tracking front-end was implemented as two threads, one for feature point management and another for feature tracking, termed here as PM and FT, respectively. The FT was implemented using the OpenCV implementation of KLT, exploiting vector processing instructions on the U3. The task of the FT thread is to track a given set of points from the previous image to the current. When the number of points being tracked falls below a given threshold, the PM initiates new and unique FAST feature in the current image. 

The EKF was implemented as two separate threads for IMU and vision measurement updates, termed here as KFI and KFV, respectively. The reason for this separation is to account for the delays in the vision pipeline. KFI thread is tasked with processing of IMU measurements, producing real-time full state estimates at 200Hz. KFV performs delayed measurement updates using a vector of corresponding point pairs provided by the FT thread and a queue of past state and state covariances stored by the KFI. A separate batch processing thread then reincorporate a stored set of IMU measurements that were obtained since the last image was captured. Batch processing thread fuses the IMU measurements in parallel to the KFI and performs an instantaneous update into the KFI when they are synchronized. This prevents the KFI from being burdened by the batch update, preserving its ability to produce real-time state estimates.

Additionally, the KFV was implemented in such a manner so that it continues to incorporate visual measurement updates from the corresponding feature point pairs, until a new set of measurements is provided by the FT. This inherently takes care of performing the maximum possible number of visual measurement updates given the amount of processing power available. The timing information for each of the threads are presented in Table \ref{tab_timing_info}. As is evident from Table \ref{tab_timing_info}, careful use of multi-threading and vector processing instructions enables the current implementation to perform visual updates at a rate close to 100Hz, although in practise it is limited by the maximum frame rate (50Hz) of the selected camera.

\begin{table}[t]
\centering
\caption{Timing information for different threads}
\begin{tabular}{ll}
\toprule
{Thread} & Average processing time\\
\midrule
KFI & 0.2ms for IMU update	\\
KFV & 2ms per update using a 40 point pairs			\\
FT & 4ms for tracking 40 points	\\
PM & 4ms for initialising 20 new features\\
\bottomrule
\end{tabular}
\label{tab_timing_info}
\end{table} 

\subsection{Results}
\begin{figure}[tb]
\centering
	\begin{subfigure}[b]{0.23\textwidth}
		\centering
		\includegraphics[width=\textwidth]{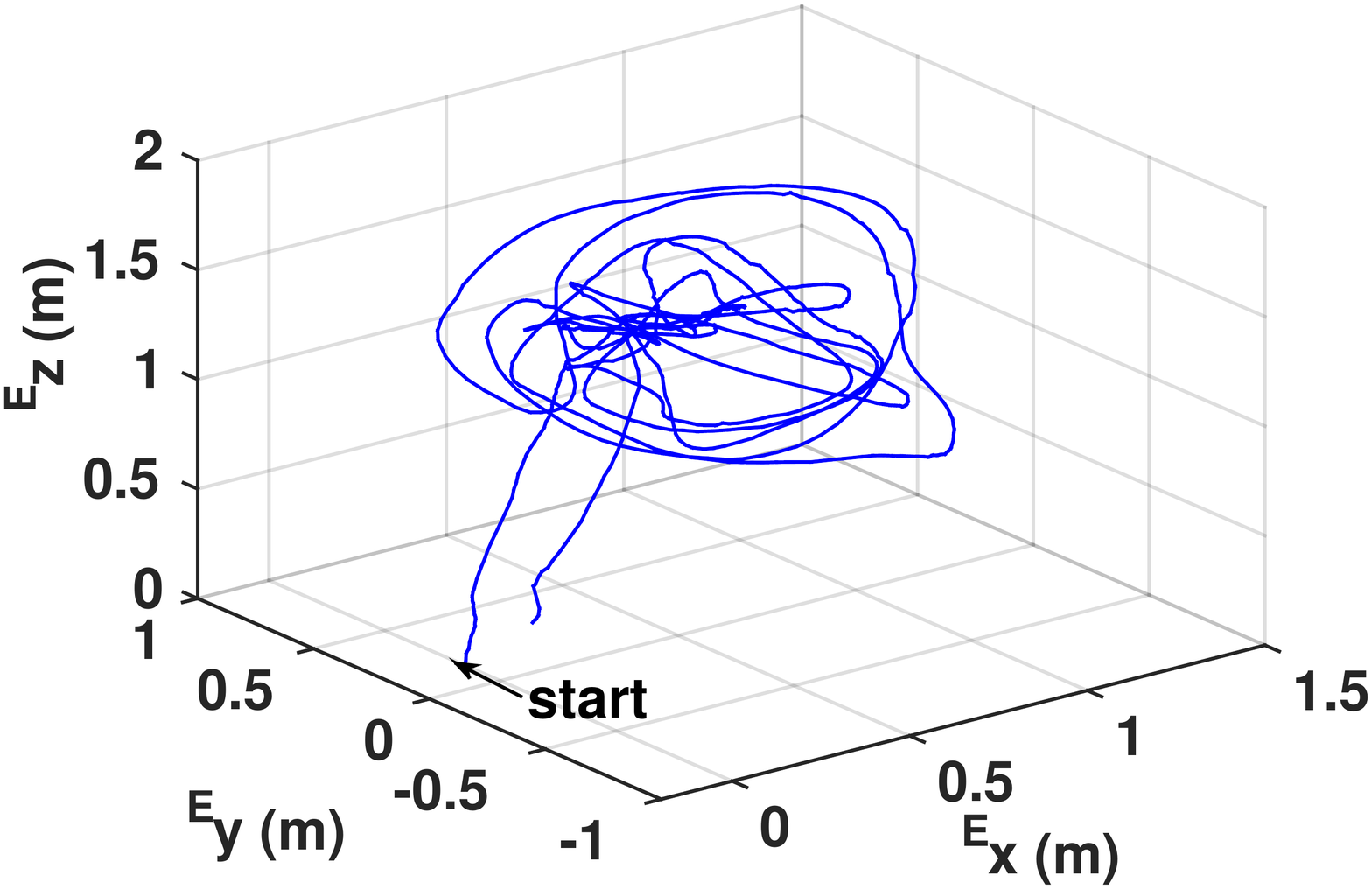}
		\caption{Circular flight path}
	\end{subfigure}
\centering
	\begin{subfigure}[b]{0.23\textwidth}
		\centering
		\includegraphics[width=\textwidth]{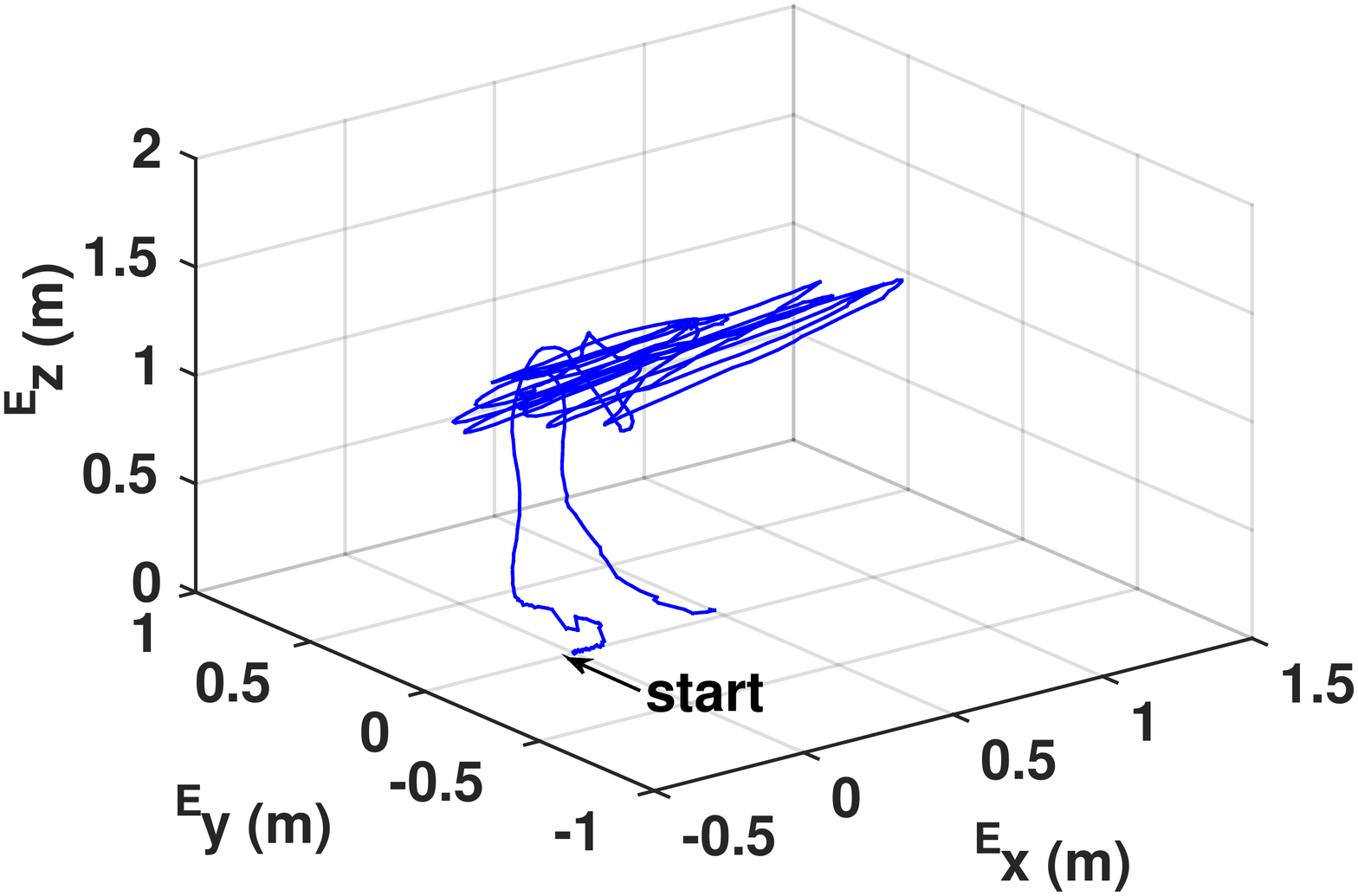}
		\caption{Linear flight path}
	\end{subfigure}
\caption[]{3D flight path of two experiments performed with the Iris+ quadrotor.}
\label{fig_3d_path}
\end{figure}

\begin{figure}[htb]
\centering
\begin{subfigure}[b]{0.23\textwidth}
\includegraphics[trim = 0cm 1.2cm 0cm 2.4cm, clip, clip, width=\textwidth]{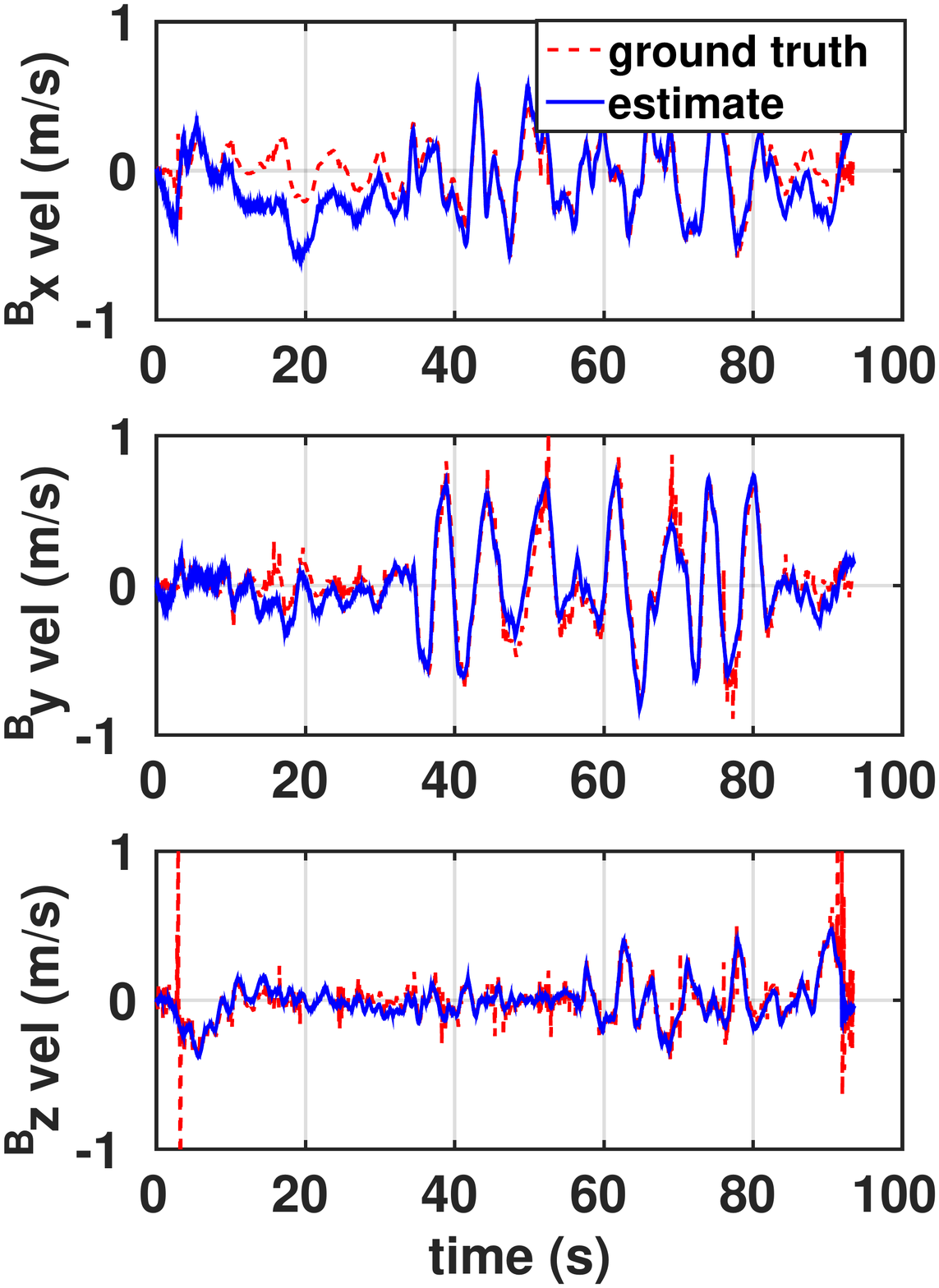}
\caption{With vision updates}
\label{fig_vel_est_exp}
\end{subfigure}
\begin{subfigure}[b]{0.23\textwidth}
\includegraphics[trim = 0cm 1.2cm 0cm 2.4cm, clip, width=\textwidth]{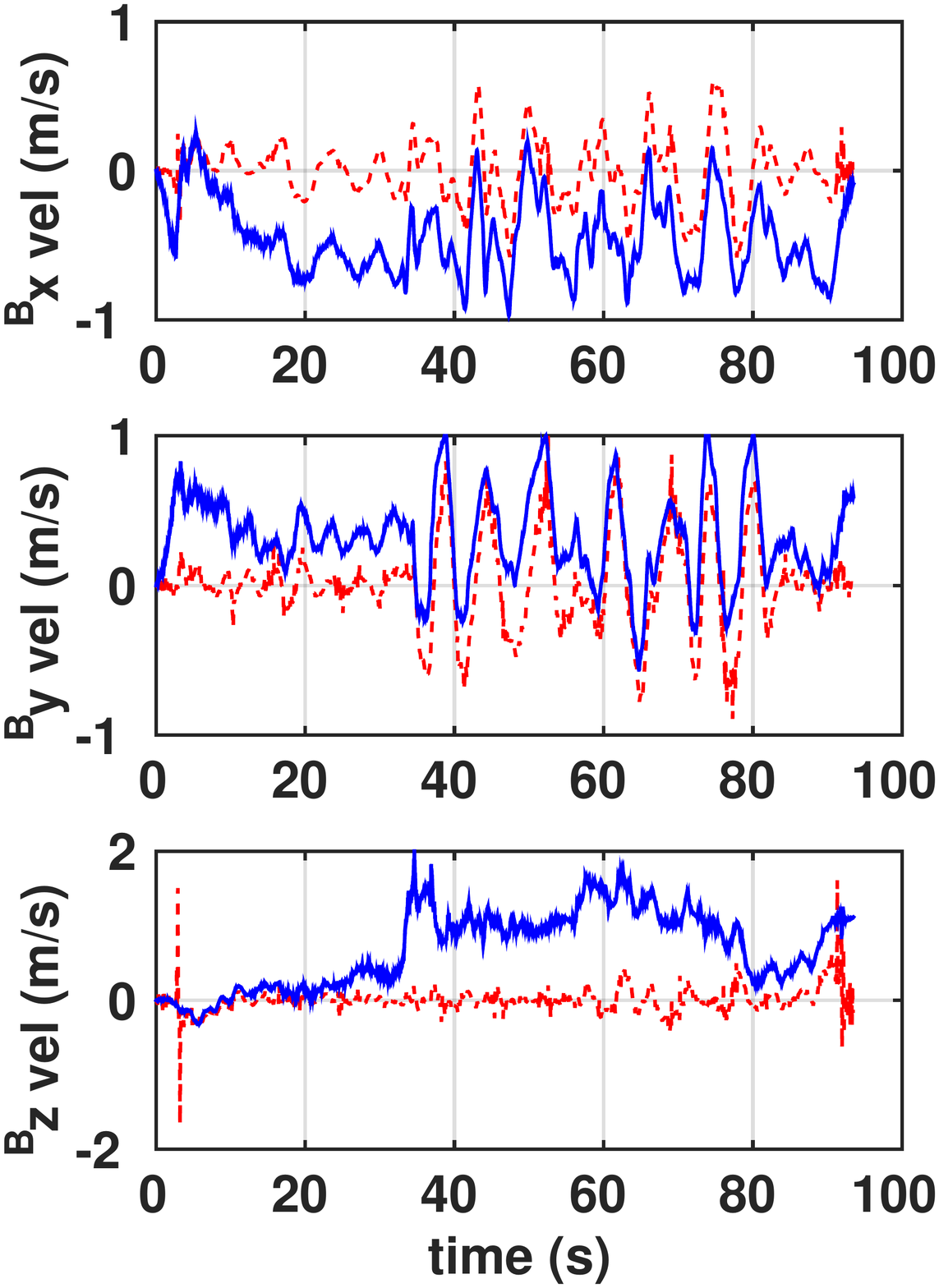}
\caption{Only inertial updates}
\label{fig_vel_est_exp_iner_only}
\end{subfigure}
\caption{Velocity estimate and ground truth for circular experimental flight with and without visual aiding. Note the scale difference in z axis.}
 \label{fig_vel_est_exp_all}
\end{figure}

\begin{figure}[htb]
\centering
\begin{subfigure}[b]{0.23\textwidth}
\includegraphics[trim = 0cm 1.2cm 0cm 2.4cm, clip, width=\textwidth]{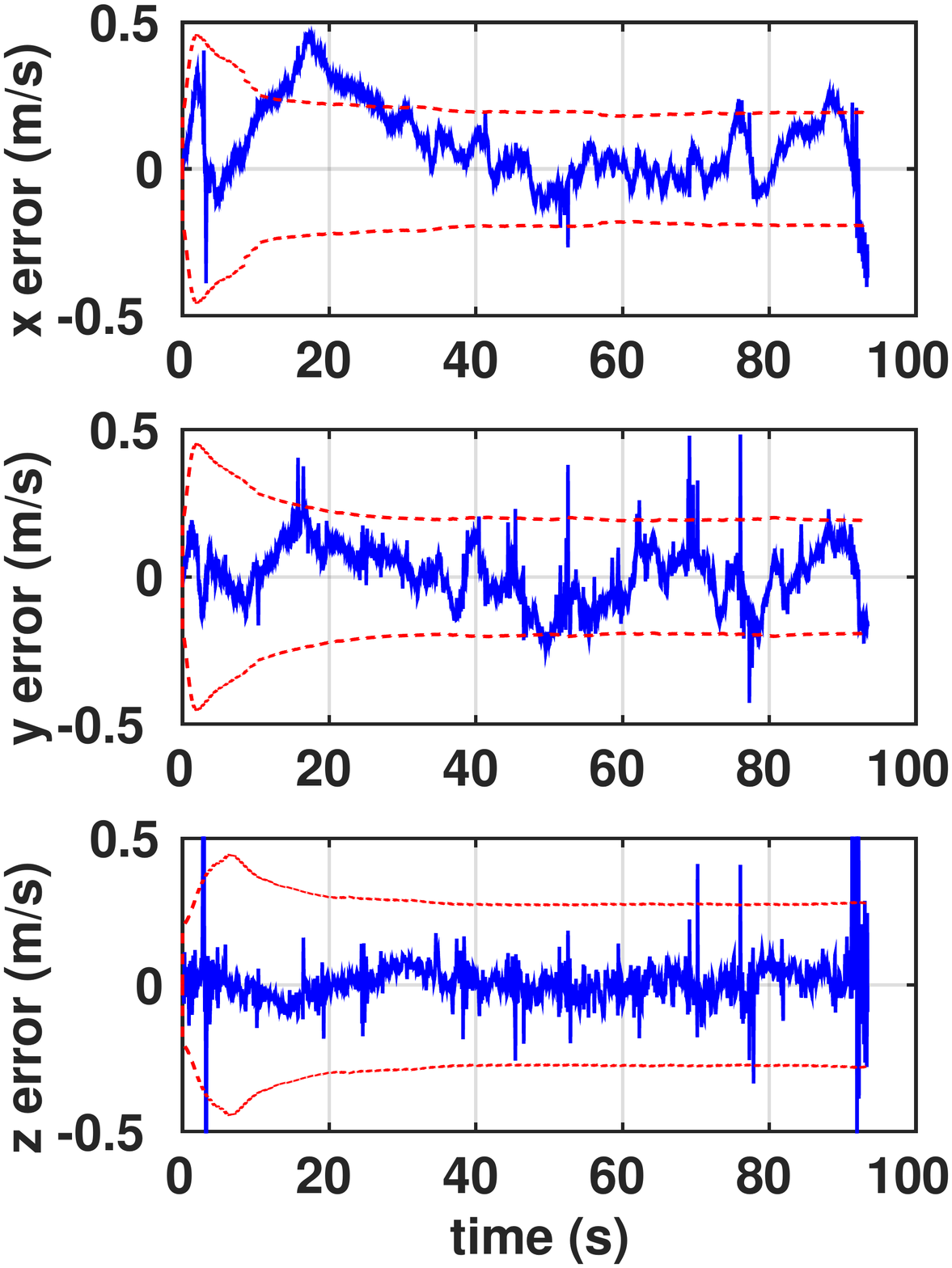}
\caption{With vision updates}
\label{fig_vel_err_exp_new}
\end{subfigure}
\begin{subfigure}[b]{0.23\textwidth}
\includegraphics[trim = 0cm 1.2cm 0cm 2.4cm, clip, width=\textwidth]{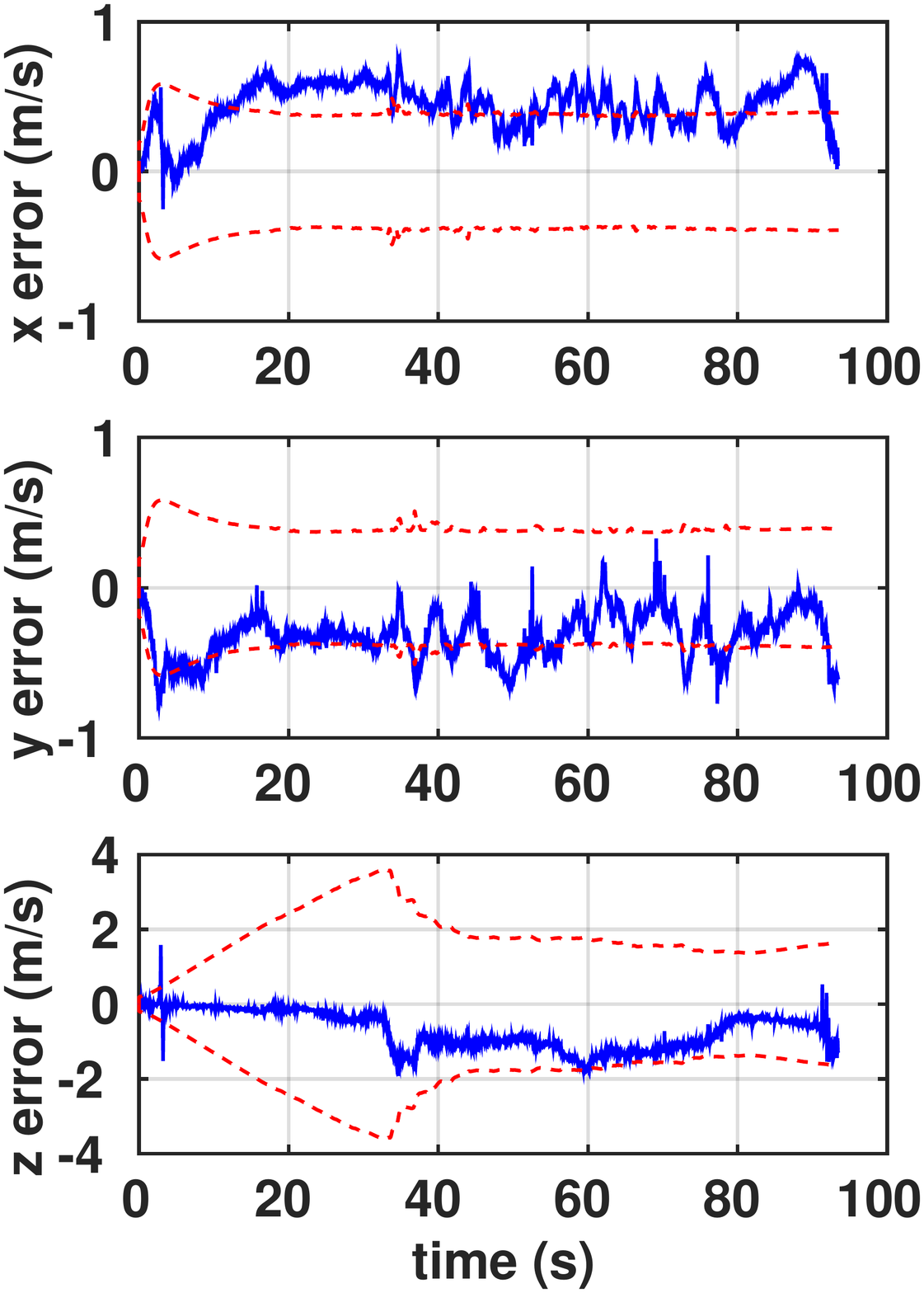}
\caption{Only inertial updates}
\label{fig_vel_err_exp_iner_only}
\end{subfigure}
\caption{Velocity estimation errors with $2\sigma$ bounds for circular experimental flight. Note the scale difference in z axis.}
 \label{fig_vel_err_exp_all}
\end{figure}

Flight tests using the Iris+ was conducted inside an indoor flight arena equipped with augmented reality tags to obtain ground truth position and orientation estimates. Fig. \ref{fig_3d_path} presents the ground truth trajectory of two such flights with different flight patterns. The velocity estimates and their respective errors of the proposed estimator for the first (Circular) flight are presented in Fig. \ref{fig_vel_est_exp_all} - \ref{fig_vel_err_exp_all}. For comparison purposes, there figures also include the velocity estimates and their errors when the visual measurement updates are disabled.

Fig. \ref{fig_vel_est_exp_all} - \ref{fig_vel_err_exp_all} demonstrate the two key improvements of the proposed design over the estimator design presented in \cite{dmw2013}. First, the visual measurement updates results in velocity estimation errors that are unbiased in $^B\bm x$ and $^B\bm y$ axes, in contrast to those with only inertial measurement updates. This improved accuracy in the proposed method stems from being able to accurately estimate the accelerometer biases. Secondly, the visual measurements also enables a considerable improvement in the estimation of the $^B\bm z$ component of the quadrotor velocity, when compared to that with only inertial measurements. Additionally, Fig. \ref{fig_vel_err_exp_all} illustrates that with the visual measurement updates, the state estimator becomes more consistent, with most of the errors contained within the $2\sigma$ bounds.  


\begin{figure}[htb]
\centering
\begin{subfigure}[b]{0.23\textwidth}
\includegraphics[trim = 0cm 1.2cm 0cm 2.4cm, clip, width=\textwidth]{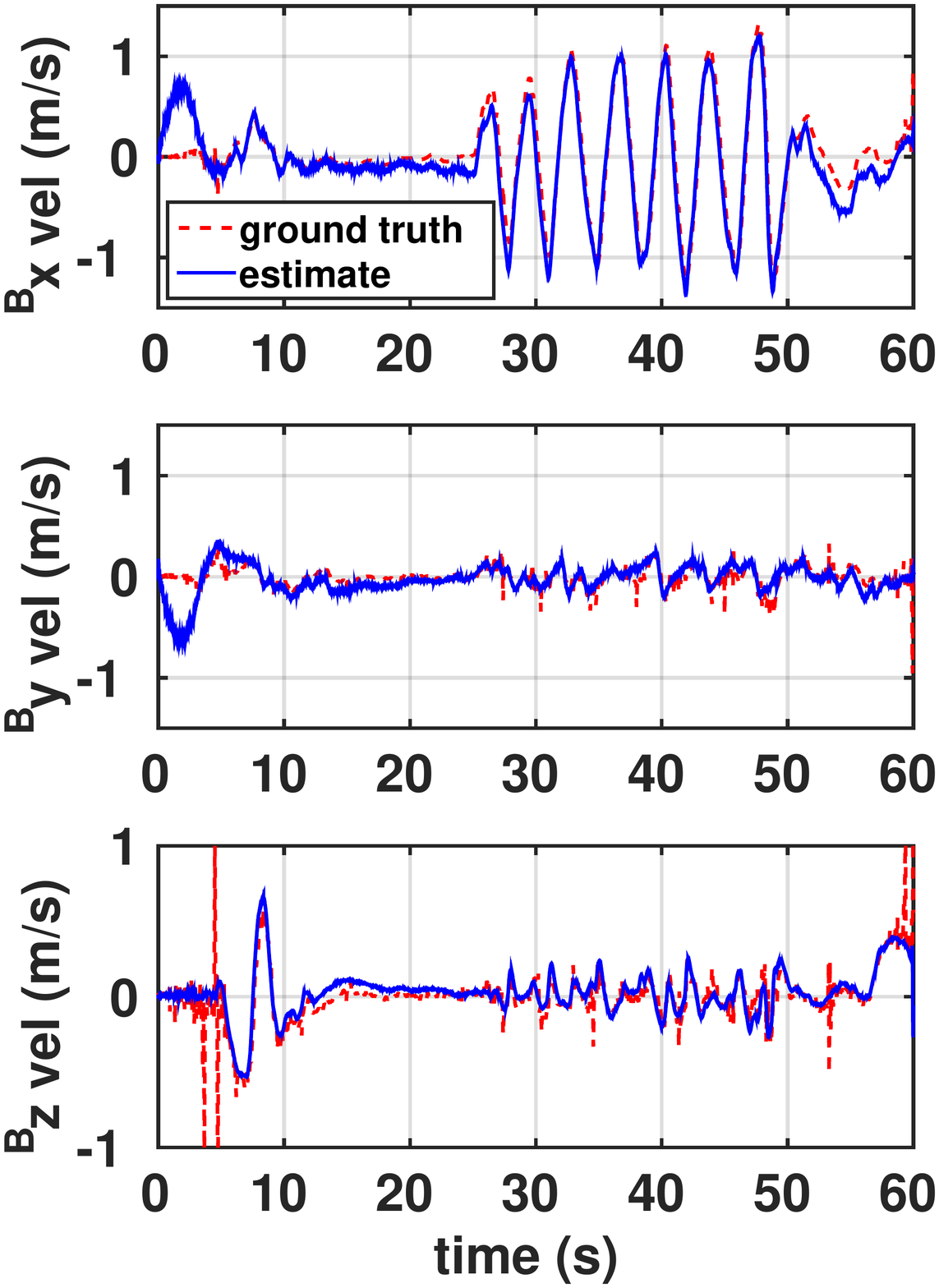}
\caption{Velocity estimates}
\label{fig_vel_est_exp_2}
\end{subfigure}
\begin{subfigure}[b]{0.23\textwidth}
\includegraphics[trim = 0cm 1.2cm 0cm 2.4cm, clip, width=\textwidth]{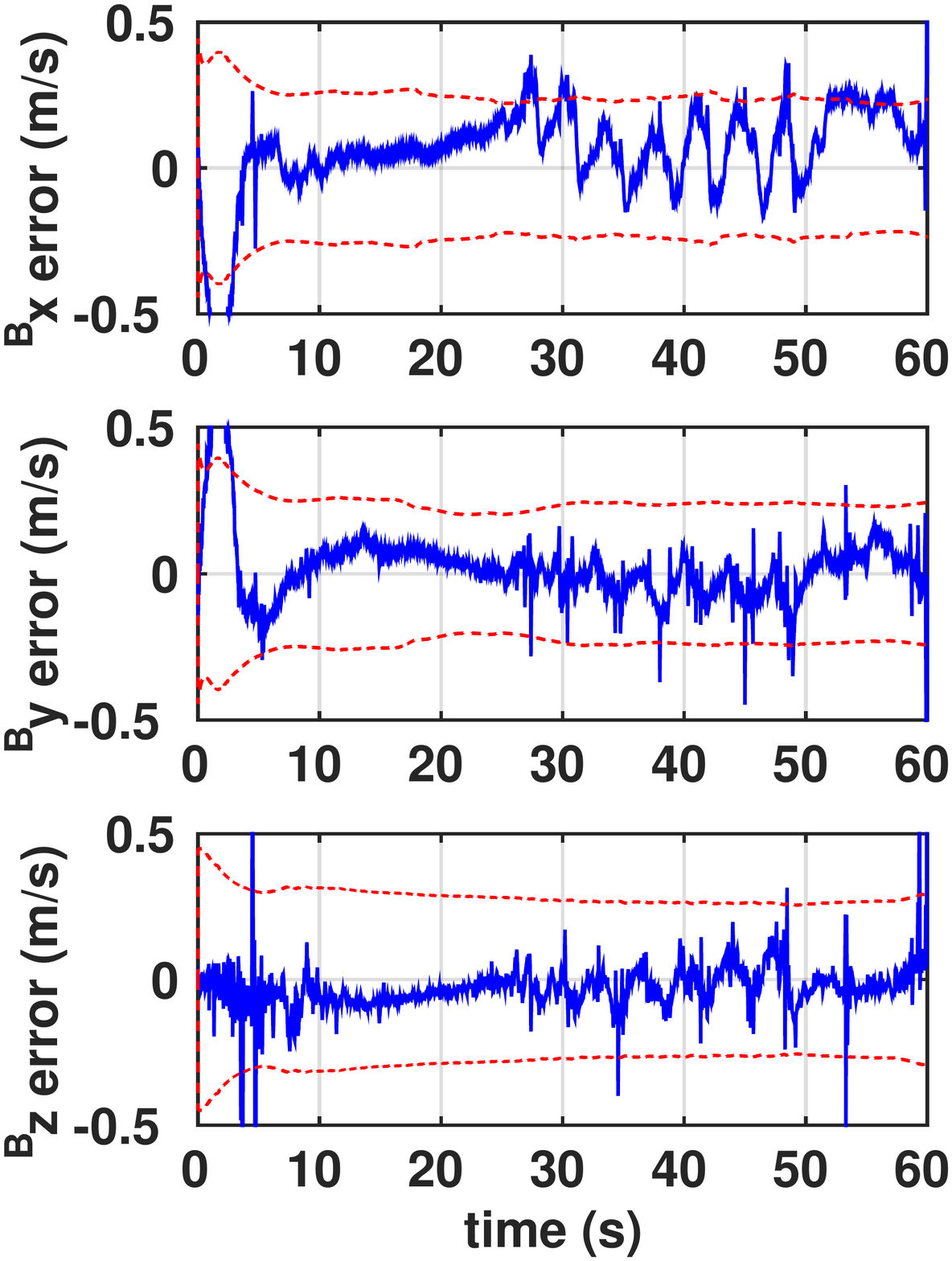}
\caption{Velocity estimation errors}
\label{fig_vel_err_exp_2}
\end{subfigure}
\caption{Velocity estimate and estimation errors with $2\sigma$ bounds for linear experimental flight.}
 \label{fig_vel_err_exp_all2}
\end{figure}

The velocity estimates and their respective errors for the second (linear) flight are presented in Fig. \ref{fig_vel_err_exp_all2}. They demonstrate that similar estimation accuracies can be obtained during hovering as well as when the quadrotor MAV motion is restricted to be mainly along the optical axis of the camera.
\section{Conclusion and Future Work}

Through simulations and experiments, this paper demonstrated that the incorporation of the visual and inertial measurements with quadrotor MAV dynamics results in a state estimator capable of producing drift free estimates of the MAV velocity in body frame as well as attitude, despite biased accelerometer and gyroscope measurements.  It also demonstrated that the utility of visual measurements reduce when the quadrotor MAV is stationary and a suitable modification to the filter design was introduced to overcome this issue with only a marginal increase in computational complexity. The ability of the estimator design to process IMU updates at 200Hz and visual updates at 50Hz in real-time on-board a 50 gram embedded computer was demonstrated via a system level implementation of the proposed algorithm. Future work focuses on coupling the estimator with a suitable velocity feedback control law to improve the autonomy of quadrotor MAVs.

\bibliographystyle{IEEEtran}
\bibliography{IEEEabrv,jreferences}

\end{document}